\documentclass[10pt, a4paper]{article}

\usepackage{lrec-coling2024} 

\usepackage{algorithm}
\usepackage{algorithmic}
\usepackage{amsmath}
\usepackage{amsfonts}
\usepackage{multirow}
\usepackage{graphicx}
\usepackage{xcolor}
\usepackage{makecell}
\usepackage{caption}

\usepackage[symbol]{footmisc}

\usepackage{newfloat}
\usepackage{listings}

\title{Enhancing Phrase Representation by Information Bottleneck Guided Text Diffusion Process for Keyphrase Extraction}

\name{
    Yuanzhen Luo\textsuperscript{\rm 1}$^\dagger$
    \thanks{$^\dagger$Work done during internship at OPPO Research Institute.},
    Qingyu Zhou\textsuperscript{\rm 2}$^\ast$\thanks{$^\ast$Corresponding author.},
    Feng Zhou\textsuperscript{\rm 2}
} 

\address {
    \textsuperscript{\rm 1}China University of Petroleum, Beijing\\
    \textsuperscript{\rm 2}OPPO Research Institute\\
    strugglingluo@gmail.com,
    qyzhgm@gmail.com, 
    zhoufeng1@oppo.com
}

\abstract{
Keyphrase extraction (KPE) is an important task in Natural Language Processing for many scenarios, which aims to extract keyphrases that are present in a given document. Many existing supervised methods treat KPE as sequential labeling, span-level classification, or generative tasks. However, these methods lack the ability to utilize the reference keyphrase information during extraction process, which may result in inferior results. In this study, we propose Diff-KPE, which leverages the supervised Variational Information Bottleneck (VIB) to guide the text diffusion process for generating enhanced keyphrase representations. Diff-KPE first generates the desired keyphrase embeddings conditioned on the entire document and then injects the generated keyphrase embeddings into each phrase representation. A ranking network and VIB are then optimized together with rank loss and classification loss, respectively. This design of Diff-KPE allows us to rank each candidate phrase by utilizing both the information of keyphrases and the document. Experiments show that Diff-KPE outperforms most of existing KPE methods on a large open domain keyphrase extraction benchmark, OpenKP, and a scientific domain dataset, KP20K.
 \\ \newline  }

\begin{document}

\maketitleabstract

\section{Introduction}

Keyphrase extraction (KPE) aims to extract several \textit{present} keyphrases from a document that can highly summarize the given document, which is helpful for many applications such as text summarization and information retrieval.

Many neural keyphrase extraction models formulate KPE as a token-level sequence labeling problem by predicting a single label for each token \cite{sahrawat2020keyphrase,alzaidy2019bi,luan2017scientific}. To use the phrase-level semantic information, some methods \cite{zhang2016keyphrase, xiong2019open, mu2020keyphrase, wang2020incorporating, sun2021capturing} modeling KPE as a phrase classification task by assigning labels to each text span. 

Different from the above methods, recent KPE models directly learn to rank each phrase \cite{mu2020keyphrase, song2021importance, sun2021capturing, song2022hyperbolic}. These methods mainly include two processes: 
candidate phrase representation construction and keyphrase importance ranking. In particular, candidate phrase representations are extracted from the token embeddings produced by pre-trained language models such as BERT \cite{devlin2018bert}, and keyphrase importance ranking usually predict a score for each candidate phrase representation and then use margin loss to sort the score of positive candidates ahead of negative ones. Since the candidate phrase representation is important for the model to score them, there are several ways to extract phrase representation: \cite{mu2020keyphrase} and \cite{sun2021capturing, song2021importance} develop Bi-LSTM and CNN to further capture the local-aware features of phrase, respectively. HyperMatch \cite{song2022hyperbolic} extract phrase representation in hyperbolic space.

Although these methods have achieved great success in many KPE benchmarks, we point out that their candidate phrase representation still lacks the utilization of reference keyphrases information. This is inspired by the intuition that how human extracts candidate phrase: They will first review the whole document and summarize a few \textit{vague} keyphrases in mind, and then take the candidate phrase and vague keyphrases both into consideration to make   the extraction decision. To achieve this process in the neural model, however, the challenge is how to generate the vague reference keyphrase information during inference time. 

To address the above issue, we propose Diff-KPE, a novel diffusion-based KPE model. We first use the diffusion model to generate a list of vague keyphrase information by recovering reference keyphrase embeddings conditioned on the whole document, then we enhance the phrase representation by injecting the vague keyphrase embeddings into each of them. To rank candidate phrases, we apply a ranking network to rank each enhanced phrase representation. By doing this, we can extract desired top $k$ keyphrases from the ranked list of phrases. In addition, we introduce a supervised Variational Information Bottleneck (VIB) to optimize a classification loss for each phrase. Supervised VIB aims to preserve the information about the target classes in the latent space while filtering out irrelevant information from the input phrase representation \cite{tishby2000information}, which helps the learning process for the vague keyphrase embedding. Multitask learning of supervised VIB can guide the model to generate informative phrase representations, thereby improving the performance of the ranking network. Overall, Diff-KPE incorporates these modules by simultaneously training these components.

Empowered by the architecture design of Diff-KPE, it exhibits the following three advantages. First, the diffusion model enables the injection of vague keyphrase information into each phrase representation, even during inference, thereby giving the model the ability to utilize keyphrase information. Second, the ranking network ranks each phrase, allowing us to flexibly extract the top k candidate keyphrases. Finally, the introduced supervised VIB guides the model to generate informative phrase representations, resulting in an improvement in ranking performance. We demonstrate the importance of each component in our experiments.
In summary, the main contributions of this paper are as follows:
\begin{itemize}
    \item We propose Diff-KPE, a diffusion-based KPE model. To the best of our knowledge, this is the first attempt to use the diffusion model for the KPE task.
    \item By incorporating the diffusion model, ranking network, and VIB into one system, we empower Diff-KPE to utilize the information of keyphrases and the document to extract candidate keyphrases.
    \item Experimental results show that Diff-KPE outperforms most of existing KPE approaches on two large keyphrase extraction benchmarks, OpenKP, and KP20K. Additionally, Diff-KPE demonstrates a more robust performance on the other five small scientific datasets.
\end{itemize}

\section{Related Work}
\subsection{Keyphrase Extraction}
Automatic KeyPhrase Extraction (KPE) aims to extract a set of important and topical phrases from a given document, which can then be used in different tasks such as summarization~\cite{li2020keywords}, problem solving~\cite{huang-etal-2017-learning,huang-etal-2018-using,huang-etal-2018-neural}, generation tasks~\cite{zhou-huang-2019-towards,li2021towards}, and so on. 

Existing KPE technologies can be categorized as unsupervised and supervised methods. Unsupervised  methods are mainly based on statistical information \cite{el2009kp, florescu2017new, campos2018text}, embedding features \cite{mahata2018key2vec, sun2020sifrank, liang2021unsupervised, zhang2021mderank, ding2021attentionrank}, and graph-based ranking algorithms \cite{mihalcea2004textrank, florescu2017positionrank, boudin2018unsupervised}. Supervised  methods commonly formulate KPE as sequence tagging approaches \cite{sahrawat2020keyphrase, alzaidy2019bi, kulkarni2022learning}, span-level classification \cite{zhang2016keyphrase,xiong2019open, mu2020keyphrase,sun2021capturing} or ranking \cite{mu2020keyphrase, song2021importance, sun2021capturing, song2022hyperbolic}, or generative tasks \cite{meng2017deep,chen2018keyphrase, yuan2018one, kulkarni2022learning}. Although supervised KPE methods require a lot of annotated data, their performance is significantly superior to unsupervised methods in many KPE benchmarks \cite{sun2021capturing, meng2017deep}. 

Recently, some works have focused on constructing a zero-shot keyphrase extractor by prompting pre-trained large language models (LLMs). For example, \cite{song2023large} verified the performance of ChatGPT \cite{chatgpt} and ChatGLM-6b \cite{zeng2022glm} for the KPE task and found that they still have a lot of room for improvement in the KPE task compared to existing SOTA supervised models. Similar results can also be observed in \cite{martinez2023chatgpt}.

\subsection{Diffusion Models for Text}
Diffusion models have been applied in many continuous domain generations like image, video, and audio \cite{kong2020diffwave, rombach2022high, ho2022imagen, yang2022diffusion}. Recently, there are some works focused on applying the diffusion model to discrete text data. They usually generate continuous representations for the desired texts/words. For example, Diffusion-LM \cite{li2022diffusion} first attempts to develop a continuous diffusion model to generate text by embedding rounding step. Following the work of Diffusion-LM, DiffuSeq \cite{gong2022diffuseq} and SeqDiffuSeq \cite{yuan2022seqdiffuseq} designed a diffusion-based sequence-to-sequence model for the text generation task. To adapt the diffusion model to longer sequence generation, \cite{zhang2023diffusum} proposed a sentence-level diffusion generation model for summary tasks, which directly generates sentence-level embeddings and matches from embeddings back to the original text. 

Contrary to previous works, we apply the diffusion model to KPE, a phrase-level extraction task. In order to take the keyphrase information into consideration during extraction, we directly inject the keyphrase information generated by the diffusion model into each phrase representation. 
\subsection{Variational Information Bottleneck in NLP}
Variational Information Bottleneck (VIB) is one of a group of Information Bottleneck (IB) methods. It aims to find compact representations of data that preserve the most relevant information while filtering out irrelevant or redundant information \cite{tishby2000information}.

There are lots of studies that apply VIB to many NLP tasks. For example, \cite{li2019specializing} used VIB for parsing, and \cite{west2019bottlesum} used it for text summarization. Recently, VIB was also used in Named Entity Recognition (NER) \cite{wang2022miner, nguyen2023span}, text classification \cite{zhang2022improving}, machine translation \cite{ormazabal2022principled} and so on.

\section{Methodology}

\begin{figure*}[ht]
    \centering
    \includegraphics[width=0.65\textwidth]{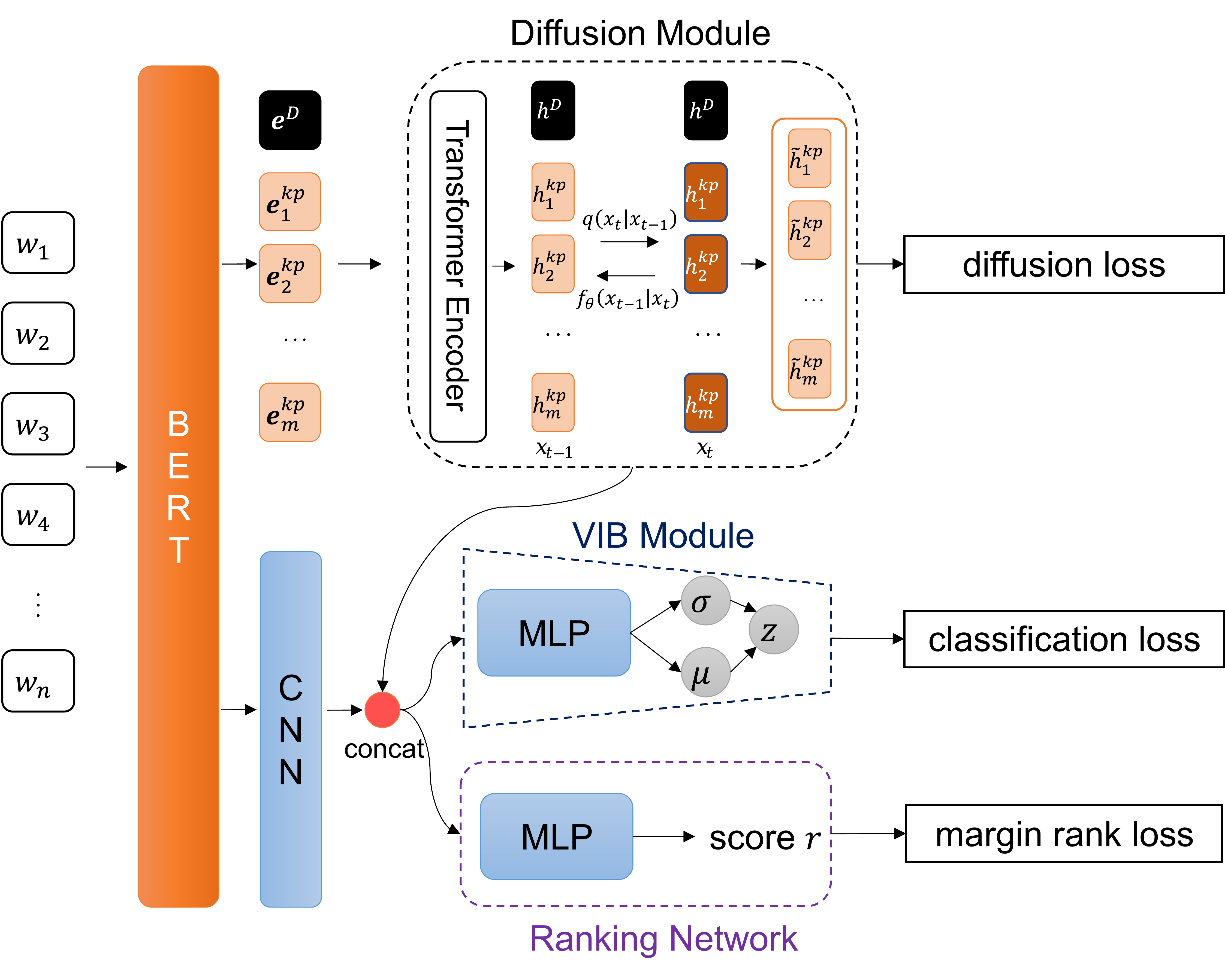}
    \caption{Diff-KPE is jointly trained with a continuous diffusion module, a variational information bottleneck, and a rank network. The black dashed box is the diffusion module, the blue dashed box is the VIB module and the purple dashed box is the rank network.}
    \label{fig:Diff-KPE}
\end{figure*}

In this section, we present the detailed design of our Keyphrase Extraction (KPE) model, named Diff-KPE. An overview of Diff-KPE is depicted in Figure \ref{fig:Diff-KPE}. Given document $D=\{w_1,w_2,...,w_n\}$, we start by enumerating all possible phrase representations and reference keyphrase embeddings. The Diffusion module is then employed to reconstruct the keyphrase embeddings and inject them into each phrase representation. Additionally, we incorporate a supervised Variational Information Bottleneck (VIB) for phrase classification and a ranking network for ranking purposes. These components work together to enhance the performance of our KPE model.

\subsection{Phrase Representation}
To enumerate and encode all the possible phrase representations, we first use pre-trained language model BERT \cite{devlin2018bert} to encode document $D=\{w_1, w_2, ..., w_n\}$,
producing contextual word embeddings $\mathbf{E}=\{\mathbf{e}_1, \mathbf{e}_2, ..., \mathbf{e}_n\}$. The word embeddings are then integrated into phrase representations using 
a set of Convolutional Neural Networks (CNNs):
\begin{equation}
  \mathbf{s}_i^k=\mathbf{CNN}^k(\mathbf{e}_i, \mathbf{e}_{i+1}, ..., \mathbf{e}_{i+n-1})
\end{equation}
where $1\leq k \leq N$ and $N$ represents the pre-defined maximum length of phrase. The $i$th k-gram phrase representation $\mathbf{s}_i^k$ is calculated by its corresponding CNN$^k$.

\subsection{Keyphrase Embeddings Generation}
In order to inject reference keyphrases information into each phrase representation, we use a continuous diffusion module to generate desired keyphrase embeddings. 
\subsubsection{Input Encoding}
To allow the diffusion module to generate desired keyphrase embeddings conditioned on the whole document, we first use another BERT model to obtain initial document and keyphrases embeddings. Refer to $m$ keyphrases and document embedding as $\mathbf{E^{kp}}=\{\mathbf{e}_i^{kp}\}_{i}^m$ and $\mathbf{e}^D$, the input encoding of the diffusion module is formatted as:
\begin{equation}
    \begin{aligned}
        \mathbf{H^{in}} &= \mathbf{h}^D || \mathbf{H}^{kp} \\
        &= \mathbf{TransfomerEncoder}(\mathbf{e}^D || \mathbf{E}^{kp})
    \end{aligned}
\end{equation}
where $\mathbf{H}^{kp}=\{\mathbf{h}_i^{kp}\}_i^m$ and $\mathbf{h}^D$ are the latent embedding of the document and $m$ keyphrases, $\mathbf{TransformerEncoder}$ is a stacked Transformer encoder which embeds the input vector into latent space, and $\mathbf{e}^D$ is the document embedding, i.e.,  the [CLS] token embedding in  BERT model, and $||$ indicates concatenation operation. Such input encoding enables our continuous diffusion
module to generate desired keyphrase embeddings conditional to the current document embeddings $\mathbf{e}^D$.

\subsubsection{Diffusion Generation Process}
Once the input encoding $\mathbf{H^{in}}$ is obtained, the diffusion model aims to perturb $\mathbf{H^{in}}$ gradually and then recover the original $\mathbf{H^{in}}$ by learning a reverse process. To achieve this, a one-step Markov transition $q(\mathbf{x}_0|\mathbf{H^{in}})$
is performed to obtain the initial state $\mathbf{x}_0$:
\begin{equation}
    \begin{aligned}
        \mathbf{x}_0&= \mathbf{x}_0^D||\mathbf{x}_0^{kp}\\
        &\sim \mathcal{N}(\mathbf{H^{in}}, \beta_0\mathbf{I})
    \end{aligned}
\end{equation}
where $\beta_t \in (0,1)$ adjusts the scale of the variance, $\mathbf{x}_0^D\sim \mathcal{N}(\mathbf{h}^D, \beta_0\mathbf{I})$ and $\mathbf{x}_0^{kp}\sim \mathcal{N}(\mathbf{H}^{kp}, \beta_0\mathbf{I})$ are the latent document embedding and keyphrase embeddings, respectively. We then start the forward process by gradually adding Gaussian noise to the latent keyphrase
embeddings $\mathbf{x}_t^{kp}$. Following the previous work \cite{zhang2023diffusum}, we keep the latent document embedding $\mathbf{x}_0^D$ unchanged, so that the
diffusion module can generate keyphrase embeddings condition to the source document. Formally, at step $t$ of the forward process $q(\mathbf{x}_t|\mathbf{x}_{t-1})$, the noised latent embedding is $\mathbf{x}_t$:
\begin{equation}
  \mathbf{x}_t = \mathbf{x}_0^D || \mathcal{N}(\mathbf{x}_t^{kp}; \sqrt{1-\beta_t}\mathbf{x}_{t-1}^{kp},\beta_t\mathbf{I})
\end{equation}
where $t\in \{1, 2, ..., T\}$ for a total of $T$ diffusion steps. For more details about the diffusion generation process, please refer to \cite{sohl2015deep}.

After adding the noise gradually at a specific time step $t$ (usually randomly choose between $[1, T]$), the backward process
is performed to recover the keyphrase embeddings $\mathbf{x}_t^{kp}$ by removing the noised. We use another stacked Transformer encoder model $f_\theta$ to conduct this backward process to recover the original input encoding $\mathbf{H}^{kp}$:
\begin{equation}
  \begin{aligned}
    \mathbf{\tilde{H}^{kp}} &= f_\theta(\mathbf{x}_t^{kp}, t)
  \end{aligned}
\end{equation}
where $f_\theta(\mathbf{x}^{kp}_t, t)$ is the stacked Transformer network to reconstruct $\mathbf{H}^{kp}$ at time step $t$.

Since the main objective of the diffusion generation module is to reconstruct the original input encoding, the objective loss
of continuous diffusion module can be defined by:
\begin{equation}
  \begin{aligned}
     \mathcal{L}_{dif} = \sum_{t=1}^{T}\|\mathbf{H}^{kp}-f_\theta(\mathbf{x}_t^{kp}, t)\|^2 + \mathcal{R}(\mathbf{x}_0)
  \end{aligned}
\end{equation}
where $\mathcal{R}(\mathbf{x}_0)$ is a regularization term for $\mathbf{x}_0$.

\subsection{Keyphrase Ranking}
After the diffusion generation process, the generated keyphrase embeddings $\tilde{\mathbf{H}}^{kp}$ are concatenated into each phrase representation $\mathbf{s}_i^k$. This aims to inject the information from keyphrases into each phrase, resulting in performance improvement of keyphrase ranking. Specifically, formulate the final phrase representation as:
\begin{equation}
  \label{eq:vib-input}
  \tilde{\mathbf{ss}}_i^k = \mathbf{s}_i^k || \mathbf{flat}(\mathbf{\tilde{H}}^{kp})
\end{equation}
where $\mathbf{flat(\mathbf{x})}$ means that $\mathbf{x}$ is flattened to a vector. Equation \ref{eq:vib-input}
means that the final phrase representation not only contains the original phrase representation but also all the 
reconstructed keyphrase information.

For training the model to rank each phrase, we introduce a contrastive rank loss. Following the previous
work \cite{sun2021capturing}, we first take a feedforward layer to project the input representation $\tilde{\mathbf{ss}}_i^k$ to a scalar score:
\begin{equation}
  r(\tilde{\mathbf{ss}}_i^k)=\mathbf{FeedForward}(\tilde{\mathbf{ss}}_i^k)
\end{equation}
Then the margin rank loss is introduced to learn to rank keyphrase $\tilde{\mathbf{ss}}_+$ ahead of non-keyphrase $\tilde{\mathbf{ss}}_-$ for the given document $D$:
\begin{equation}
  \mathcal{L}_{rank} = \sum_{\tilde{\mathbf{ss}}_+, \tilde{\mathbf{ss}}_-\in D} \max(0, 1-r(\tilde{\mathbf{ss}}_+)+r(\tilde{\mathbf{ss}}_-))
\end{equation}

\subsection{Keyphrase Classification}
Combining the keyphrase classification task during training can enhance the phraseness measurement of the phrase \cite{sun2021capturing, song2021importance}. Similar to previous work \cite{xiong2019open, sun2021capturing, song2021importance}, 
we introduce a classification loss for each final phrase representation for multi-task learning. We found that the use of supervised VIB substantially improves the ranking performance (See Ablation Study). 
Supervised VIB aims to preserve the information about the target classes in the latent space while filtering out irrelevant
information from the input \cite{voloshynovskiy2019information}. Given the final phrase representation $\tilde{\mathbf{ss}}_i^k$, the supervised VIB first compresses the input to a latent variable $z\sim q_{\phi_1}(z|\tilde{\mathbf{ss}}_i^k)$. We apply two linear layers to construct the parameters $q$ using the following equations:
\begin{equation}
    \begin{aligned}
        \mu&=\mathbf{W}_\mu \tilde{\mathbf{ss}}_i^k + \mathbf{b}_\mu \\
        \sigma^2&=\mathbf{W}_\sigma \tilde{\mathbf{ss}}_i^k + \mathbf{b}_\sigma
    \end{aligned}
\end{equation}
where $\mu$ and $\sigma$ are the parameters of a multivariate Gaussian, representing the latent feature space of the phrase; $\mathbf{W}$ and $\mathbf{b}$ are weights and biases of the linear layer, respectively. The posterior distribution $z\sim q_{\phi_1}(z|\tilde{\mathbf{ss}}_i^k)$ is approximated via reparameterisation trick \cite{kingma2013auto}:
\begin{equation}
    z=\mu+\sigma\epsilon,\text{where } \epsilon\sim \mathcal{N}(0, 1)
\end{equation}
Since the main objective of VIB is to preserve target class information while filtering out irrelevant information from the input, the objective loss function for the supervised VIB is based on classification loss and compression loss. Denoted by $y$ as the true label of the input phrase, the objective loss of the supervised VIB is defined as:
\begin{equation}
  \begin{aligned}
    \mathcal{L}_{vib}(\phi) &= \mathbb{E}_{z}[-\log p_{\phi_2}(y|z)] \\
    &+ \alpha \mathbb{E}_{\tilde{\mathbf{ss}}_i^{k}}[D_{KL}(q_{\phi_1}(z|\tilde{\mathbf{ss}}_i^{k}),pr(z))]
  \end{aligned}
  \label{eq:vib-loss}
\end{equation}
where $pr(z)$ is an estimate of the prior probability $q_{\phi_1}(z)$, $\alpha$ range in $[0,1]$, $\phi$ is the neural network parameters, and $D_{KL}$ is the Kullback-Leibler divergence. We use a multi-layer perceptron with one linear layer and softmax function to calculate $p_{\phi_2}(y|z)$. Note that Equation \ref{eq:vib-loss} can be approximated by the Monte Carlo sampling method with sample size $M$.

\subsection{Optimization and Inference}
We jointly optimize the diffusion module, ranking network, and supervised VIB end-to-end. Specifically, the overall training objective loss can 
be represented as:
\begin{equation}
  \mathcal{L} = \mathcal{L}_{dif} + \mathcal{L}_{vib} + \mathcal{L}_{rank}
\end{equation}
For inference, the Transformer encoder first obtains the initial document embeddings $\mathbf{h}^D$, and then the one-step Markov $q(\mathbf{x}_0^D|\mathbf{h}^D)$
is performed. To construct the noise keyphrase embedding $\mathbf{x}_T^{kp}$, we random sample $m$ Gaussian noise embeddings
such that $\mathbf{x}_T^{kp}\sim \mathcal{N}(\mathbf{0}, \mathbf{I})$. Then the reverse process is applied to remove the Gaussian
noise of $\mathbf{x}_T=\mathbf{x}_0^D || \mathbf{x}_T^{kp}$ iteratively and get the output keyphrase embeddings 
$\mathbf{\tilde{H}}^{kp}=[\tilde{\mathbf{h}}^{kp}_1, \tilde{\mathbf{h}}^{kp}_2, ..., \tilde{\mathbf{h}}^{kp}_m]$. After that, each original phrase representation $\mathbf{s}_i^{k}$ is concatenated to the flattened keyphrase embeddings $\mathbf{\tilde{H}}^{kp}$ and input to the ranking network to obtain the final score for each phrase.

\section{Experiments}

\subsection{Datasets}
In this paper, we use seven KPE benchmark datasets in our experiments.

\begin{itemize}
    
    \item \textbf{OpenKP} \cite{xiong2019open} consists of around 150K web documents from the Bing search engine. We follow its official split of training (134K), development (6.6K), and testing (6.6K) sets. Each document in OpenKP was labeled with 1-3 keyphrases by expert annotators.

    \item \textbf{KP20K} \cite{meng2017deep} consists of a large amount of high-quality scientific metadata in the computer science domain from various online digital libraries \cite{meng2017deep}. We follow the original partition of training (528K), development (20K), and testing (20K) set.

    \item \textbf{SemEval-2010} \cite{kim2013automatic} contains 244 scientific documents. The official split of 100 testing documents is used for testing in our experiments. 

    \item \textbf{SemEval-2017} \cite{augenstein2017semeval} contains 400 scientific documents. The official split of 100 testing documents isis used for testing in our experiments.

    \item \textbf{Nus} \cite{nguyen2007keyphrase} contains 211 scholarly documents. We treat all 211 documents as testing data.

    \item \textbf{Inspec} \cite{hulth2003improved} contains 2000 paper abstracts. We use the original 500 testing papers and their corresponding controlled (extractive) keyphrases for testing.

    \item \textbf{Krapivin} \cite{krapivin2009large} contains 2305 papers from scientific papers in ACM. We treat all 2305 papers as testing data.
\end{itemize}

\begin{table}[ht]
    \centering
    \resizebox{\linewidth}{!}{
    \begin{tabular}{ccccc}
        \hline
        \textbf{Dataset} & \makecell[c]{\textbf{Avg.} \\ \textbf{Doc Len.}} & \makecell[c]{\textbf{Avg.}\\ \textbf{KP Len.}} & \makecell[c]{\textbf{Avg.}\\ \textbf{\# KP}} & \makecell[c]{\textbf{\% up to}\\ \textbf{5-gram}} \\
        \hline
        OpenKP & 1212.3 & 2.0 & 2.2 & 99.2\% \\
        KP20k & 169.3 & 1.9 & 3.5 & 99.8\% \\
        \hline
        SemEval-2010 & 9664.2 & 2.0 & 9.5 & 99.8\%\\
        SemEval-2017 & 190.6 & 2.3 & 11.3 & 97.9\%\\
        Nus & 8707.4 & 1.9 & 8.0 & 99.8\% \\
        Inspec & 138.9 & 2.2 & 6.4 & 99.8\%\\
        Krapivin & 9354.1 & 1.9 & 3.8 & 99.9\%\\
        \hline
    \end{tabular}
    }
    \caption{
    Statistics of benchmark datasets, including the average length of the document, the average length of the keyphrase, the average number of extractive keyphrases, and the percentage of keyphrases with a maximum length of 5.
    }
    \label{tab:stat}
\end{table}

Note that in order to verify the robustness of our model, we test the model trained with KP20K on the testing data of SemEval-2010, SemEval-2017, Nus, Inspec, and Krapivin. For all datasets, only the \textit{present} keyphrases are used for training and testing. The statistics of the training set of OpenKP and KP20k are shown in Table \ref{tab:stat}.

\subsection{Baselines}
To keep consistent with previous work \cite{meng2017deep, xiong2019open, mu2020keyphrase, sun2021capturing}, we compare our model with two categories of KPE methods: Traditional KPE baselines and Neural KPE baselines. 

Traditional KPE baselines consist of two popular unsupervised KPE methods, statistical feature-based method TF-IDF \cite{sparck1972statistical} and graph-based method TextRank \cite{mihalcea2004textrank}, and two feature-based KPE systems PROD \cite{xiong2019open} and Maui \cite{medelyan2009human}.

\begin{table*}[ht]
\small
    \centering
    \begin{tabular}{c|cc|cc|cc|cc|c}
        \hline
        \multirow{2}*{\textbf{Model}} & \multicolumn{6}{|c|}{\textbf{OpenKP}} & \multicolumn{2}{|c|}{\textbf{KP20k}} &
        \multirow{2}*{\textbf{Averge}}
        \\
        \cline{2-9}
        ~ & \multicolumn{2}{|c|}{F1@1\ R@1} & \multicolumn{2}{|c|}{F1@3\ R@3} & \multicolumn{2}{|c|}{F1@5\ R@5} & \multicolumn{2}{|c|}{F1@5\ F1@10} & ~
        \\
        \hline
        \multicolumn{10}{l}{\textbf{Traditional KPE}} \\
        \hline
        TF-IDF$^\dagger$\cite{sun2021capturing} & 19.6 & 15 & 22.3 & 28.4 & 19.6 & 34.7 & 10.8 & 13.4 & 20.4 \\
        TextRank$^\dagger$ \cite{sun2021capturing} & 5.4 & 4.1&7.6&9.8&7.9&14.2&18.0&15.0 & 10.2 \\
        Maui$^\dagger$ \cite{mu2020keyphrase} & - & - & - & - & - & - & 27.3 & 24.0 & - \\
        PROD$^\dagger$ \cite{xiong2019open} & 24.5 & 18.8 & 23.6 & 29.9 & 18.8 & 33.1 & - & - & - \\
        \hline
        \multicolumn{10}{l}{\textbf{Neural KPE}} \\
        \hline
        CopyRNN$^\dagger$ \cite{meng2017deep} & 21.7&17.4&23.7&33.1&21&41.3&32.7&27.8 & 27.3\\
        BLING-KPE$^\dagger$ \cite{xiong2019open} & 28.5&22.0& 30.3&39.0&27.0&48.1&-&- & - \\
        SKE-Base-Cls$^\dagger$ \cite{mu2020keyphrase}& - & - & - & - & - & - & 39.2 & 33.0 & - \\
        BERT-Span$^*$ \cite{sun2021capturing} & 34.1 & 28.9&34.0&49.2&29.3&59.3&39.3&32.5 &38.3\\
        BERT-SeqTag$^*$ \cite{sun2021capturing} &37.0&31.5&37.4&54.1&31.8&64.2&40.7&33.5 & 41.2\\
        ChunkKPE$^*$ \cite{sun2021capturing}&37.0&31.4&37.0&53.3&31.1&62.7&41.2&33.7 & 40.9\\
        RankKPE$^*$ \cite{sun2021capturing}&36.9&31.5&38.1&55.1&32.5&65.5&41.3&34.0 & 41.8 \\
        JointKPE$^*$ \cite{sun2021capturing}& 37.2 &  31.8 & 38.2 & 55.2 & 32.6 & \textbf{65.7} & 41.1 & 33.8 & \underline{41.9} \\
        JointKPE$^\dagger$ \cite{sun2021capturing}& 37.1 &  31.5 & 38.4 & 55.5 & 32.6 & \textbf{65.7} & 41.1 & 33.8 & \underline{41.9}\\
        KIEMP$^\dagger$ \cite{song2021importance} & 36.9 & 29.8 & \textbf{39.2} & 51.7 & \textbf{34.0} & 61.5 & \textbf{42.1} & \textbf{34.5} & 41.2\\ 
        HyperMatch$^\dagger$ \cite{song2022hyperbolic} & 36.4 & 29.5 & 39.0 & 51.5 & 33.7 & 61.2 & 41.6 & 34.3 & 40.9 \\
        \hline
        \multicolumn{10}{l}{\textbf{LLM (zero shot)}} \\
        \hline
        ChatGLM2-6b$^\dagger$ \cite{song2023large} & 16.0 & - & 11.0 & - & 8.6 & - & - & - & -\\
        GPT-3.5-turbo$^*$ \cite{chatgpt} & 20.8 & 17.0 & 20.4 & 26.9 & 16.6 & 30.0 & 13.5 & 10.8 & 19.5 \\
        \hline
        \textbf{Diff-KPE} & \textbf{37.8} & \textbf{32.2} & 38.5 & \textbf{55.6} & 32.7 & \textbf{65.7} & 41.7 & 34.3 & \textbf{42.3}\\
        \hline
    \end{tabular}

    \caption{Overall performance of 
    extractive KPE models on OpenKP development set and KP20k testing set. \textbf{Bold} indicates the best results, and \underline{underlined} are the SOTA baselines.
    $^\dagger$ indicates results are copied from corresponding papers, and $^*$ are from our reproduction. Note that HyperMatch and KIEMP use the RoBERTa \cite{liu2019roberta} as backbone, while others use BERT-base  \cite{devlin2018bert}. } 
    \label{tab:openkp-kp20k-res}
\end{table*}

\begin{table*}[ht]
\small
    \centering
    \begin{tabular}{c|cc|cc|cc|cc|cc|c}
        \hline
        \multirow{2}*{\textbf{Model}} & \multicolumn{2}{|c|}{\textbf{SemEval-2010}} & \multicolumn{2}{|c|}{\textbf{SemEval-2017}} & \multicolumn{2}{|c|}{\textbf{Nus}}& \multicolumn{2}{|c|}{\textbf{Inspec}} & \multicolumn{2}{|c|}{\textbf{Krapivin}} &
        \multirow{2}*{\textbf{Average}}
        \\
        \cline{2-11}
        ~ & \multicolumn{2}{|c|}{F1@5\ F1@10} & \multicolumn{2}{|c|}{F1@5\ F1@10} & \multicolumn{2}{|c|}{F1@5\ F1@10} & \multicolumn{2}{|c|}{F1@5\ F1@10} & \multicolumn{2}{|c|}{F1@5\ F1@10} & ~ \\
        \hline
        TF-IDF & 12.0&18.4&-&-&13.9&18.1&22.3&30.4&11.3&14.3& -\\
        TextRank & 17.2&18.1&-&-&19.5&19.0&22.9&27.5&17.2&14.7 & -\\
        JointKPE$^*$ & 28.2 & 31.0 & 29.6 & \textbf{37.7} & 33.9 & 35.0 & 31.8 & 35.0 & 33.3 & 29.2& 32.4\\
        ChatGLM2-6b$^\dagger$ & 13.2 & 13.8 & - & - & - & - & 25.1 & 30.1 & - & - & -\\
        GPT-3.5-turbo$^*$ & 11.2 & 11.1 & 17.0 & 25.8 & 14.8 & 15.2 & 30.6 & 33.9 & 12.8 & 11.7 & 18.4\\
        \textbf{Diff-KPE} & \textbf{29.3} & 31.0 & \textbf{29.7} & 37.2 & \textbf{35.2} & \textbf{36.0} & \textbf{32.3} & 35.0 & \textbf{35.0} & \textbf{31.4} & \textbf{33.2}\\
        \hline
    \end{tabular}
    \caption{
    Evaluation results on five small 
    scientific testing sets. The results are evaluated using the models trained on KP20k. Bold indicates the best results. $^*$ results are obtained from our reproduction.
    }
    \label{tab:other-res}
\end{table*}

Neural KPE baselines consist of a sequence-to-sequence generation-based model named CopyRNN \cite{meng2017deep}. Previous state-of-the-art method on OpenKP and KP20K, KIEMP \cite{song2021importance} incorporating multiple perspectives estimation for phrase ranking. Another strong baseline, JointKPE \cite{sun2021capturing}, including its two variants ChunkKPE and RankKPE are reproduced according to their open-source code\footnote{https://github.com/thunlp/BERT-KPE}. HyperMatch \cite{song2022hyperbolic}, a new matching method for extracting keyphrase in the hyperbolic space. Two phrase-level classification-based models named SKE-Base-Cls \cite{mu2020keyphrase} and BLING-KPE \cite{xiong2019open}. We also compare our model with BERT-based span extraction and sequence tagging methods, both of which come from the implementation of \cite{sun2021capturing}. Note that since both of KeyBart and KBIR  \cite{kulkarni2022learning} are pre-trained with a well-defined pretraining strategy specifically on RoBERTa-large \cite{liu2019roberta}, we do not compare our Diff-KPE with them for a fair comparison.

In addition, we also add the results of two large language models (LLMs) with zero-shot settings: ChatGLM2-6b \cite{zeng2022glm} and ChatGPT\footnote{GPT-3.5-turbo, version: gpt-3.5-turbo-0125} \cite{chatgpt}. To restrict the output format of ChatGPT, we design the following prompt template:

\texttt{[Instruction]}

\texttt{Please extract 1 to 15 keyphrases from the given document. Your extracted keyphrases should reasonably represent the topic of the document and must appear in the original text. You must give the keyphrases by strictly following this format: ``[extracted keyphrases]", for example: ``[machine learning, neural networks, NLP]"}

\texttt{[Document]}

\texttt{\{document\}}

It should be noted that designing more complex prompts may improve the performance of LLMs, which is beyond the scope of this paper.

\subsection{Evaluation Metrics}
We use Recall (R), and F-measure (F1) of the top $K$ predicted keyphrases for evaluating the performance of the KPE models. Following the prior research \cite{meng2017deep, xiong2019open}, we utilize $K=\{1, 3, 5\}$ on OpenKP and $K=\{5, 10\}$ on others. When determining the exact match of keyphrases, we first lowercase the candidate keyphrases and reference keyphrases, and then we apply Porter Stemmer \cite{porter1980algorithm} to both of them.

\subsection{Implementation details}
We truncate or zero-pad each document due to the input length limitations (512 tokens). We use the base version of BERT to generate initial word embeddings. We also use the base version of Sentence-BERT \cite{reimers2019sentence} to generate initial fixed phrase embeddings for the diffusion module. The maximum length of k-gram is set to $N=5$ for all datasets. The maximum diffusion time steps $T$ is set to 100, $\alpha=2.8e-6$. The hidden size and number of layer in Transformer encoder in the diffusion module are set to 8 and 6 respectively. The latent dimension of the VIB model is set to 128. Sample size $M=5$. We optimize Diff-KPE using AdamW with 5e-5 learning rate, 0.1 warm-up proportion, and 32 batch size.  The training used 8 NVIDIA Tesla V100 GPUs and took about 20 hours on 5 epochs. During training Diff-KPE, we also set a simple early stop strategy such that the model would stop training if the validation performance (F1@3 for OpenKP, F1@5 for KP20K) does not improve after 5 times consecutive evaluations (We evaluate the model every 200 optimization steps), and we select the model with the best validation performance. We run our model with 5 different random seeds and report their average score.

\begin{table*}[htbp]
    \centering
    \begin{tabular}{c|ccc|ccc|ccc}
        \hline
        \textbf{Setting}&\multicolumn{3}{|c|}{F1@1 P@1 R@1}&\multicolumn{3}{|c|}{F1@3 P@3 R@3}&\multicolumn{3}{|c}{F1@5 P@5 R@5} \\
        \hline
         
         \textbf{Diff-KPE} & \textbf{37.8} & \textbf{51.4} & \textbf{32.2} & \textbf{38.5} & \textbf{31.4} & \textbf{55.6} & \textbf{32.7} & \textbf{22.8} & \textbf{65.7} \\
           - \textit{w/o} VIB & 36.5&49.4&31.09&37.7&30.8&54.5&32.1&22.4&64.8 \\
           - \textit{w/o} diffusion& 36.6&49.7&31.2	&37.9&31.0&54.8	&32.3&22.5&65.0 \\
         \hline
    \end{tabular}
    \caption{Evaluation metrics on the OpenKP development set by different settings. ``\textit{w/o} VIB'' means Diff-KPE without VIB module, ``\textit{w/o} diffusion'' means Diff-KPE without diffusion module.}
    \label{tab:ablation}
\end{table*}

\section{Results and Analysis}

In this section, we present the evaluation results of the proposed Diff-KPE on seven widely-used benchmark datasets (OpenKP, KP20k, SemEval-2010, SemEval-2017, Nus, Inspec, Krapivin).

\subsection{Overall Performance}
Table \ref{tab:openkp-kp20k-res} shows the evaluation results of Diff-KPE and baselines.
Based on the results, it is evident that the neural KPE methods outperform all the traditional KPE algorithms. Among the traditional methods, the unsupervised methods TF-IDF and TextRank show stable performance on both OpenKP and KP20k datasets, while the feature-based methods PROD and Maui outperform them on OpenKP and KP20k respectively. This is not surprising, as supervised methods benefit from large annotated data during training.

For neural KPE methods, CopyRNN performs the worst as it also focuses on generating abstractive keyphrases. HyperMatch, JointKPE and its variant RankKPE show powerful performance, outperforming other baselines such as the phrase classification-based models BLING-KPE, SKE-Base-Cls, BERT-Span, and the sequence tagging method BERT-SeqTag. It is worth noting that BERT-SeqTag and ChunkKPE exhibit competitive performance compared to RankKPE, indicating their robustness and strong performance.

Overall, Diff-KPE outperforms all baselines excluding KIEMP on both OpenKP and KP20K datasets. Compared to JointKPE, Diff-KPE shows slight improvements in F1@3 and F1@5 but a dramatic improvement in F1@1. Compared to the previous SOTA neural baseline method KIEMP, KIEMP outperforms our Diff-KPE in most F1@k scores on OpenKP and KP20k. However, Diff-KPE still exhibits performance improvements in F1@1, R@1, R@3, and R@5 on OpenKP. We hypothesize that the improvements in Recall benefit from our diffusion module, which is able to inject generated keyphrase embeddings into phrase representations, thereby enhancing the recall performance.

Moreover, to verify the robustness of Diff-KPE, we also evaluate our model trained with the KP20k dataset on five additional small scientific datasets, as shown in Table \ref{tab:other-res} \footnote{We cannot evaluate KIEMP due to lack of open source code and models \cite{song2021importance}.}. Diff-KPE demonstrates better or competitive results on all datasets compared to the best baseline JointKPE. We believe this phenomenon arises from the benefit of the diffusion module: during inference, the diffusion model can generate candidate keyphrase embeddings, providing keyphrase information for the ranking network to better rank each phrase.

\subsection{Ablation Study}
\label{sec:ablation}
To understand the effect of each component on our Diff-KPE model. We perform the ablation study on the OpenKP development set as following settings:
\begin{itemize}
    \item - \textit{w/o} VIB: replace the VIB model with a single feedforward layer for keyphrase classification.
    \item - \textit{w/o} diffusion: the diffusion model is removed, and only use the phrase representations obtained from CNNs for ranking and classifying.
    \item Diff-KPE: the original full joint model.
\end{itemize}
As shown in Table \ref{tab:ablation}, the absence of the diffusion model or VIB model results in a dramatic drop in performance across all metrics, particularly in F1@1 (1.2 and 1.3 respectively). This performance decline indicates the crucial role of both the diffusion and VIB models in keyphrase ranking. The strong performance of Diff-KPE can be attributed to two main advantages.
Firstly, the diffusion module directly incorporates the semantic information of keyphrases into the final phrase representations.
Secondly, the supervised VIB module introduces an external classification loss during training, which indirectly enhances the diffusion module or CNNs to generate more informative n-gram embeddings.
Therefore, it is evident that the addition of the diffusion module and supervised VIB greatly contributes to the overall performance improvement.

\begin{table*}[ht]
\small
    \centering
    \begin{tabular}{|l|}
        \hline
         \textbf{(1) Partial Document}: \\
         ... in Comics RealWorld Objects Non canon \textcolor{red}{Adventure Time} comic English Share Adventure Time is a comic \\ book series published by BOOM Studios written by Dinosaur Comics creator \textcolor{red}{Ryan North}, and illustrated by \\ Shelli Paroline and Braden Lamb. The comic book is released monthly beginning with issue 1 in February 2012 \\ ... (URL: http:\/\/adventuretime.wikia.com\/wiki\/Adventure\_Time\_(comic)) \\
         \textbf{Reference Keyphrases}: \\
         \textcolor{red}{adventure time}; \textcolor{red}{ryan north} \\
         \hline
         \textbf{\textit{Without diffusion module}}: \\
         \textcolor{red}{adventure time}; boom studios; comic book series; dinosaur comics; \textcolor{red}{ryan north} \\
         \textbf{\textit{Diff-KPE}}: \\
         \textcolor{red}{adventure time}; comic book series; \textcolor{red}{ryan north}; comic book; dinosaur comics \\
        \hline
        \hline
        \textbf{(2) Partial Document}: \\
        \textcolor{red}{CodeSnip}: How to Run Any \textcolor{red}{Oracle Script} File Through Shell Script in UNIX ... by Deepankar Sarangi ... Listing \\ 1 ... The first line is a comment line which is UNIX kernel specific. In the following approach the available shell is \\ KORN shell ... \\ (URL: http://aspalliance.com/1589\_CodeSnip\_How\_to\_Run\_Any\_Oracle\_Script\_File\_Through\_Shell\_Script\_in\_ \\ UNIX.4)
        \textbf{Reference Keyphrases}: \\
        \textcolor{red}{codesnip}; \textcolor{red}{oracle script} \\
        \hline
        \textbf{\textit{Without diffusion module}}: \\
        shell script; oracle script file through; oracle script file; shell scripts; \textcolor{red}{codesnip} \\
        \textbf{\textit{Diff-KPE}}: \\
        unix; shell script; \textcolor{red}{codesnip}; oracle script file through; \textcolor{red}{oracle script} \\
        \hline
    \end{tabular}
    \caption{Example of keyphrase extraction results on two selected OpenKP development examples. The phrase in red is the desired reference keyphrase. }
    \label{tab:case-study}
\end{table*}

\section{Case Study}
To further demonstrate the effectiveness of the diffusion module in Diff-KPE, we provide examples of the extracted keyphrases from our different models (Diff-KPE and Diff-KPE without diffusion module). Two typical cases from the development set of OpenKP are shown in Table \ref{tab:case-study}.

In case (1), both Diff-KPE and Diff-KPE without diffusion successfully extract the desired reference keyphrases ``adventure time" and ``ryan north" within their top 5 ranked prediction phrases. However, Diff-KPE ranks the phrase ``ryan north" higher, resulting in a higher F1@3 score in this case. This illustrates that adding the diffusion module helps the desired keyphrase representation obtain a higher rank score.

Similarly, in case (2), Diff-KPE ranks the desired keyphrases ``codesnip" and ``oracle script" higher compared to the model without diffusion. As a result, Diff-KPE successfully extracts all the reference keyphrases in case (2). The main reason for these results may be that the keyphrase embeddings generated by the diffusion module are directly injected into each phrase representation, enabling the ranking network to better rank each phrase by utilizing the keyphrase information.

\begin{figure}[h]
    \centering
    \includegraphics[width=0.95\linewidth]{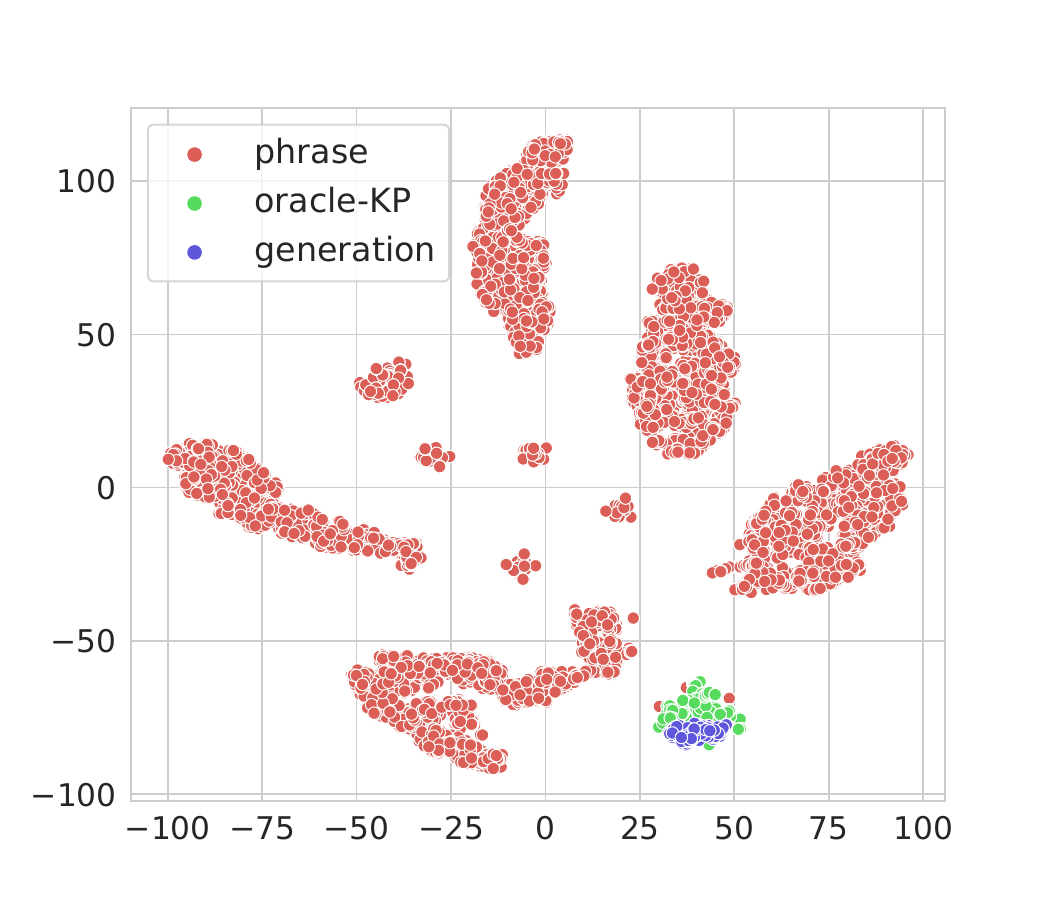}
    \caption{T-SNE visualization of phrase embeddings from OpenKP dataset.}
    \label{fig:tsne}
\end{figure}

We also analyze the generated keyphrase embeddings quality. We apply T-SNE \cite{van2008visualizing} to reduce all the phrase representation's dimensions to 2 in Figure \ref{fig:tsne}. We can find that the oracle keyphrases (green dots) and generated keyphrases (blue dots) are clustered together and far away from most non-keyphrase embeddings (red dots). This finding demonstrates that our diffusion model is powerful in recovering keyphrase embeddings.

\section{Conclusion}
In this paper, we propose Diff-KPE, a novel joint keyphrase extraction (KPE) model composed of three essential modules: the diffusion module, the ranking network, and a supervised VIB module. Each component plays a crucial role in learning expressive phrase representations. The diffusion module is responsible for generating keyphrase embeddings, effectively infusing keyphrase semantic information into the final phrase representation. Simultaneously, the supervised VIB introduces a classification loss for each phrase, encouraging the model to generate more informative representations and ultimately improving the ranking performance.Experimental results on seven keyphrase extraction benchmark datasets demonstrate the effectiveness and superiority of Diff-KPE.

However, since our model requires many steps of forward noise injection and backward denoising, our Diff-KPE is about 2x slower than the previous SOTA model JointKPE during inference. Moreover, our model also lacks the ability to generate abstractive keyphrases. In future work, we plan to improve the computation efficiency and explore the application of Diff-KPE in abstractive keyphrase generation, leveraging its powerful architecture and flexibility for generating concise and informative keyphrases.

\section{Ethics Statement}
We take ethical considerations seriously and strictly adhere to the Ethics Policy. This paper focuses on the attempt to the application of diffusion model for keyphrase extraction. Both the datasets and base models used in this paper are publicly available and have been widely adopted by researchers. We ensure that the findings and conclusions of this paper are reported accurately and objectively.

\bibliographystyle{lrec-coling2024-natbib}
\bibliography{lrec-coling2024-example}

\bibliographystylelanguageresource{lrec-coling2024-natbib}
\bibliographylanguageresource{languageresource}

\end{document}